\documentclass[sigconf]{acmart}

\usepackage{amsthm}
\usepackage{multirow}
\usepackage{stfloats}
\usepackage{graphicx}
\usepackage{url}

\usepackage{subcaption}

\AtBeginDocument{%
  }

\copyrightyear{2023}
\acmYear{2023}
\setcopyright{acmlicensed}\acmConference[CIKM '23]{Proceedings of the 32nd ACM International Conference on Information and Knowledge Management}{October 21--25, 2023}{Birmingham, United Kingdom}
\acmBooktitle{Proceedings of the 32nd ACM International Conference on Information and Knowledge Management (CIKM '23), October 21--25, 2023, Birmingham, United Kingdom}
\acmPrice{15.00}
\acmDOI{10.1145/3583780.3615160}
\acmISBN{979-8-4007-0124-5/23/10}

\begin{document}

\setlength{\abovedisplayskip}{3pt} % adjust the value as desired
\setlength{\belowdisplayskip}{3pt} % adjust the value as desired

\title{STAEformer: Spatio-Temporal Adaptive Embedding Makes Vanilla Transformer SOTA for Traffic Forecasting}
% Staeformer, STAEformer, Trafformer, STAEM (model), STAEN (network), Staemer, STAE, STAEmb
% Spatio-Temporal Adaptive Embedding 是整体的模型名还是就是E_a? 很关键
% 模型名里面不蹭former 比较安全
% Spatio-Temporal Adaptive Embedding Is All You Need for Traffic Transformer
% Spatio-Temporal Adaptive Embedding Is All You Need for Traffic Forecasting (Pass)

\author{Hangchen Liu*\textsuperscript{1}, Zheng Dong*\textsuperscript{1}, Renhe Jiang\textsuperscript{\textdagger2}, Jiewen Deng\textsuperscript{1}, \\Jinliang Deng\textsuperscript{3}, Quanjun Chen\textsuperscript{2}, Xuan Song\textsuperscript{\textdagger1,2}}

\thanks{* Equal contribution.}
\thanks{\textdagger~Corresponding author.}

\affiliation{
\textit{\textsuperscript{1}Southern University of Science and Technology}, \textit{\textsuperscript{2}The University of Tokyo}, 
\textit{\textsuperscript{3}University of Technology Sydney}
\country{}
}

% \email{{liuhc3, zhengdong00, dengjw1}@outlook.com, {jiangrh, songxuan}@csis.u-tokyo.ac.jp,}
% \email{jinliang.deng@student.uts.edu.au, chen1990@iis.u-tokyo.ac.jp}

\email{{liuhc3, zhengdong00}@outlook.com, jiangrh@csis.u-tokyo.ac.jp}

\renewcommand{\shortauthors}{H. Liu, Z. Dong, and R. Jiang et al.}

% \settopmatter{authorsperrow=3}

% \author{Hangchen Liu} \authornote{Equal contribution.}
% \author{Zheng Dong} \authornotemark[1]
% % \email{{liuhc3, zhengdong00}@outlook.com}
% \affiliation{
%   \institution{Southern University of Science and Technology, China}
%   \streetaddress{}
%   \city{}
%   \state{}
%   \country{}
%   \postcode{}
% }

% \author{Renhe Jiang}
% % \email{jiangrh@csis.u-tokyo.ac.jp}
% \authornote{Corresponding Author.}
% \affiliation{
%   \institution{The University of Tokyo,}
%   \streetaddress{}
%   \city{}
%   \state{}
%   \country{Japan}
%   \postcode{}
% }

% \author{Jiewen Deng} %\email{dengjw1@outlook.com}
% \affiliation{   \institution{Southern University of Science and Technology, China}   \streetaddress{}   \city{}   \state{}   \country{}   \postcode{} }

% \author{Jinliang Deng} %\email{jinliang.deng@student.uts.edu.au} 
% \affiliation{   \institution{University of Technology Sydney, Australia}   \streetaddress{}   \city{}   \state{}   \country{}   \postcode{} }

% \author{Quanjun Chen} %\email{chen1990@iis.u-tokyo.ac.jp} 
% \affiliation{   \institution{The University of Tokyo,}   \streetaddress{}   \city{}   \state{}   \country{Japan}   \postcode{} }

% \author{Xuan Song} %\email{songx@sustech.edu.cn}
% \affiliation{   \institution{Southern University of Science and Technology, China}   \streetaddress{}   \city{}   \state{}   \country{}   \postcode{} }

\begin{abstract}
With the rapid development of the Intelligent Transportation System (ITS), accurate traffic forecasting has emerged as a critical challenge. The key bottleneck lies in capturing the intricate spatio-temporal traffic patterns. In recent years, numerous neural networks with complicated architectures have been proposed to address this issue. However, the advancements in network architectures have encountered diminishing performance gains. In this study, we present a novel component called \textbf{\textit{spatio-temporal adaptive embedding}} that can yield outstanding results with vanilla transformers. Our proposed \underline{S}patio-\underline{T}emporal \underline{A}daptive \underline{E}mbedding trans\underline{former} (\textbf{STAEformer}) achieves state-of-the-art performance on six real-world traffic forecasting datasets. Further experiments demonstrate that spatio-temporal adaptive embedding plays a crucial role in traffic forecasting by effectively capturing intrinsic spatio-temporal relations and chronological information in traffic time series.

\end{abstract}


% \ccsdesc[300]{Information systems~Data mining}
% \ccsdesc[500]{Computing methodologies~Knowledge representation and reasoning}


\maketitle

\begin{figure}[!t]
    \centering
    \begin{subfigure}[t]{0.192\linewidth}
        \centering
    \includegraphics[width=\linewidth]{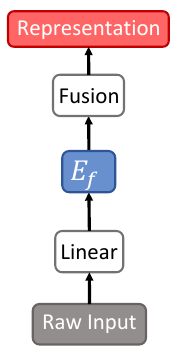}
        \caption{STGNNs}
        \label{fig: embedding_STGNNs}
    \end{subfigure}
        \begin{subfigure}[t]{0.30\linewidth}
        \centering
        \includegraphics[width=\linewidth]{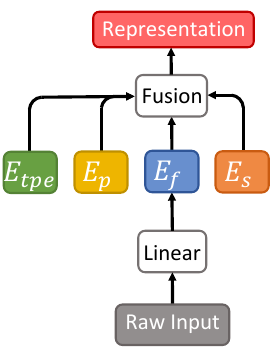}
        \caption{Transformer-\\based Models}
        \label{fig: embedding_PDFormer}
    \end{subfigure}
    \begin{subfigure}[t]{0.23\linewidth}
        \centering
        \includegraphics[width=\linewidth]{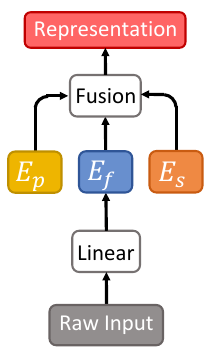}
        \caption{Spatial Temporal Identity}
        \label{fig: embedding_STID}
    \end{subfigure}
    \label{fig: embeddings}
    \begin{subfigure}[t]{0.23\linewidth}
        \centering
        \includegraphics[width=\linewidth]{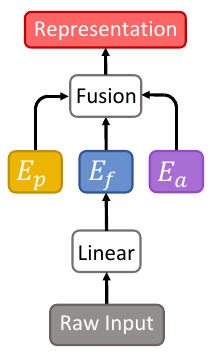}
        \caption{Ours}
        \label{fig: embedding_ours}
    \end{subfigure}
    \caption{Input Embeddings in Different Traffic Models.}
    \label{fig: model_embeddings}
\end{figure}

% 而且我们自己先说为啥没加别的former 说的时候俩原因 1. 别的没有特化在traffic 任务 2. 一些共通的embedding 可以对应于哪一种图一的哪一种embedding  我们的文章不是去提新的 transformer
\section{Introduction}
% With the development of Intelligent Transportation System(ITS), traffic prediction becomes one of the most crucial problems.
Traffic forecasting~\cite{jiang2021dl} aims to predict the future traffic time series in road networks based on historical observations. % Given its significant role in many real-world applications, it's remained an enduring research topic in both academia and industry.
In recent years, the success of deep learning models has been notable, primarily attributable to their ability to capture the inherent spatio-temporal dependencies in the traffic system. Among them, Spatio-Temporal Graph Neural Networks (STGNNs)~\cite{STGCN, DCRNN} and Transformer-based models~\cite{ASTGCN, GMAN, STTN, PDFormer} become very popular for their outstanding performance. Researchers have devoted considerable effort to developing fancy and complicated models for traffic forecasting, such as novel graph convolutions~\cite{StemGNN, STSGCN, STFGNN, STMGCN, CCRNN, DMSTGCN, STGNN, TGCN, Z-GCNETs, HGCN, AutoSTGCN, STGODE, STAGGCN}, learning graph structures~\cite{GWNet, MegaCRN, GTS, MTGNN, SLCNN}, efficient attention mechanisms~\cite{STWA, Informer, Autoformer, Pyraformer, FEDformer}, and other methods~\cite{STMetaNet, D2STGNN, PMMemNet, EnhanceNet, AutoSTG, STEP}. Nonetheless, the advancements in network architectures have encountered diminishing performance gains, prompting a shift in focus from complicated model designs towards effective representation techniques for the data itself. 

% , we emphasize the importance of the simple yet powerful \textbf{\textit{input embedding}}.

In light of this, in this study, we focus on \textbf{\textit{input embedding}}, a widely-used, simple, yet powerful representation technique, that is often overlooked by many researchers in terms of its effectiveness. Specifically, it adds an embedding layer on the input, providing multiple types of embeddings for the model backbone. Figure~\ref{fig: model_embeddings} presents a comparative analysis of the utilized embeddings by previous models. STGNNs mainly use feature embedding $E_f$, i.e. a transformation to project the raw input into hidden space. Transformer-based models require additional knowledge such as temporal positional encoding $E_{tpe}$ and periodicity (daily, weekly, monthly) embedding $E_p$ due to the attention mechanism's inability to preserve the positional information of the time series. Recent models, including PDFormer~\cite{PDFormer}, GMAN~\cite{GMAN} and STID~\cite{STID}, apply spatial embedding $E_s$. Notably, STID~\cite{STID} is among the few studies that explore these embeddings. It employs spatial embedding and temporal periodicity embedding with a simple Multi-layer Perceptron (MLP) and achieves remarkable performance.

% To further improve the effectiveness of embedding, 
To further enhance the representation effectiveness, 
we propose a novel \textbf{\textit{spatio-temporal adaptive embedding}} $E_a$ and apply it on the vanilla Transformer~\cite{attention} along with $E_p$ and $E_f$, which is shown in Figure~\ref{fig: embedding_ours}. In detail, the raw input goes through an embedding layer to obtain the input embedding, which is fed into temporal and spatial transformer layers followed by a regression layer to make predictions. Our proposed model, named \underline{S}patio-\underline{T}emporal \underline{A}daptive \underline{E}mbedding trans\underline{former} (\textbf{STAEformer}), has a far more concise architecture but achieves the state-of-the-art (SOTA) performance. In our model, $E_a$ plays a crucial role by effectively capturing intrinsic spatio-temporal relations and chronological information in traffic time series. Experiments and analyses on six real-world traffic datasets prove that our proposed $E_a$ can make vanilla transformer SOTA for traffic forecasting.

% \vspace{-0.3cm}
\section{Problem Definition}
Given the traffic series $X_{t-T+1:t}$ in the previous $T$ time frames, traffic forecasting aims to infer the traffic data in the future $T'$ frames by training a model $\mathbb{F(\cdot)}$ with parameters $\theta$, which can be formulated as:
\begin{equation}
\begin{aligned}
[X_{t-T+1},\dots,X_{t}] \xrightarrow[\theta]{\mathbb{F(\cdot)}} [X_{t+1},\dots,X_{t+T'}]
\end{aligned}
\label{eq:problem}
\end{equation}
\noindent where each frame $X_i\in \mathbb{R} ^{N\times d}$, $N$ is the number of spatial nodes. $d$ is the dimension of the input feature which equals 1 in our case, standing for traffic volume. 
% For accurate description, we introduce particles to represent the entities on temporal and spatial axis. Specifically, there are $T\times N$ particles in traffic series data with $T$ frames and $N$ spatial nodes. 

% \begin{figure}[!t]
%     \centering
%     \label{fig: embeddings}
%     \begin{subfigure}[t]{0.55\linewidth}
%         \centering
%         \includegraphics[width=\linewidth]{figure/Transformer Positional Enbedding.pdf}
%         \caption{Vanilla Transformer Positional Embedding}
%         \label{fig: tpe}
%     \end{subfigure}
%     \begin{subfigure}[t]{0.44\linewidth}
%         \centering
%         \includegraphics[width=\linewidth]{figure/Spatial Embedding.pdf}
%         \caption{Spatial Embedding}
%         \label{fig: se}
%     \end{subfigure}
%     \label{fig: embeddings}
%     \begin{subfigure}[t]{0.6\linewidth}
%         \centering
%         \includegraphics[width=\linewidth]{figure/Learnable Temporal Embedding.pdf}
%         \caption{Learnable Temporal Embedding}
%         \label{fig: lte}
%     \end{subfigure}
%     \begin{subfigure}[t]{0.39\linewidth}
%         \centering
%         \includegraphics[width=\linewidth]{figure/Spatio-Temporal Adaptive Embedding.pdf}
%         \caption{Spatio-Temporal Adaptive Embedding}
%         \label{fig: adp}
%     \end{subfigure}
%     \caption{Different Spatial and Temporal Embeddings in traffic forecasting.}
% \end{figure}

\begin{figure*}[t]   \centering \includegraphics[width=175mm]{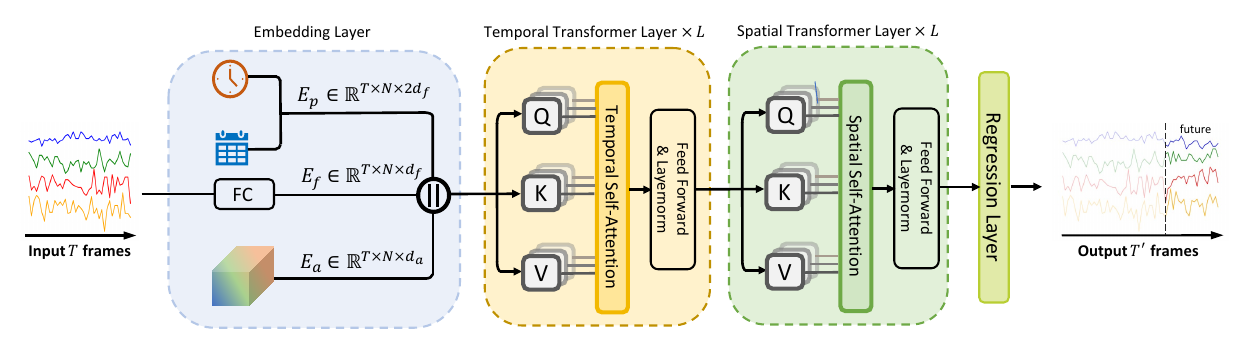} %\caption{Our Proposed Embedding and Vanilla Spatial/Temporal Transformer}   
\caption{The Architecture of \underline{S}patio-\underline{T}emporal \underline{A}daptive \underline{E}mbedding trans\underline{former} (\textbf{STAEformer}).}
\label{fig:STAdp}   \end{figure*}

% \vspace{-0.3cm}
\section{Methodology}
As shown in Figure~\ref{fig:STAdp}, our model consists of an embedding layer, vanilla transformers applied along temporal axis as the temporal transformer layer and along spatial axis as the spatial transformer layer, then a regression layer. The embedding layer obtains the hidden representation fused by the feature embedding, the periodicity embedding, and the spatio-temporal adaptive embedding. The spatio-temporal transformer layers capture traffic relations, followed by a regression layer making the prediction.

\subsection{Embedding Layer}
To keep the native information in the raw data, we utilize a fully connected layer to obtain the feature embedding ${E_f} \in \mathbb{R} ^{T\times N \times d_f }$:
\begin{equation}
\begin{aligned}
E_f = FC(X_{t-T+1:t})
\end{aligned}
\label{eq:ife}
\end{equation}
\noindent where $d_f$ is the dimension of the feature embedding, and $FC(\cdot)$ indicates a fully connected layer.

% \begin{figure}[t]   \centering \includegraphics[width=50mm]{figure/Architecture.pdf} \caption{The Architecture of STAdp}   \label{fig:architecture}   \end{figure}

% \begin{figure*}[t]   \centering \includegraphics[width=175mm]{figure/STAdp.pdf} 	\caption{The Architecture of STAdp}   \label{fig:STAdp}   \end{figure*}

% TODO 这一段感觉写的有点复杂了，其实可以直接语言描述一下：Tw 和 Td 是两个embedding dict，然后根据下标获取对应 embs
% We denote the learnable day-of-week matrix as $T_w\in \mathbb{R} ^{N_w\times d_f}$, and the learnable timestamps-of-day matrix as $T_d\in \mathbb{R} ^{N_d\times d_f}$, where $N_w$ is the number of days in a week, and $N_d$ is the number of timestamps in each day. Then, the day-of-week embedding $E_{w(t)} \in \mathbb{R}^{d_f}$, the timestamp-of-day embedding $E_{d(t)} \in \mathbb{R}^{d_f}$, and the periodicity embedding $E_{p(t)} \in \mathbb{R}^{2d_f}$ at time frame $t$ can be formulated as:
% \begin{equation}
% \left\{
% \begin{aligned}
% E_{w(t)} &= query(T_w,W_t)\\
% E_{d(t)} &= query(T_d,D_t)\\
% E_{p(t)} &= (E_{w(t)}||E_{d(t)})
% \end{aligned}
% \right.
% \end{equation}
% \noindent where $W_t \in \mathbb{R}^T$, $D_t \in \mathbb{R}^T$ denote the day-of-week data and the timestamp-of-day data for the traffic series data in $t-T:t$. Here, $||$ denotes the concatenation operation, and the operation $query$ is formulated as:
% \begin{equation}
% \begin{aligned}
% query(A,B)=\{\ A_i|A_i = A_{B_i},\ i=(0,\cdots,T)\}.
% \end{aligned}
% \label{eq:query}
% \end{equation}
% Thus, the periodicity embedding for the raw data $E_p \in \mathbb{R}^{T\times N \times 2d_f}$.

We denote the learnable day-of-week embedding dict as $T_w\in \mathbb{R} ^{N_w\times d_f}$, and the timestamps-of-day embedding dict as $T_d\in \mathbb{R} ^{N_d\times d_f}$, where $N_w$ = 7 denotes the number of days in a week. In our case, $N_d$ = 288 is the number of timestamps in each day. Denoting $W^t \in \mathbb{R}^T$, $D^t \in \mathbb{R}^T$ as the day-of-week data and the timestamp-of-day data for the traffic time series in $t$-$T$+1:$t$, we use them as indices to extract the corresponding day-of-week embedding $E_{w}^{t} \in \mathbb{R}^{T\times d_f}$ and timestamp-of-day embedding $E^{t}_d \in \mathbb{R}^{T\times d_f}$ from the embedding dicts. By concatenating and broadcasting them, we get the periodicity embedding $E_p \in \mathbb{R}^{T\times N \times 2d_f}$ for the traffic time series.

On one hand, it's intuitive that the temporal relation is not only decided by periodicity but also affected by the chronological order in the traffic time series. For example, a time frame in traffic time series should be more similar to the time frames nearby. On the other hand, the time series from different sensors tend to have different temporal patterns. Thus, instead of adopting a pre-defined or dynamic adjacency matrix for spatial relation modeling, we designed a spatio-temporal adaptive embedding $E_a \in \mathbb{R} ^{T \times N \times d_a}$ to capture intricate spatio-temporal relation in a uniform way. In particular, $E_a$ is shared across different traffic time series.

By concatenating the embeddings above, we obtain the hidden spatio-temporal representation $Z \in \mathbb{R}^{T \times N \times d_h}$ as follows:
\begin{equation}
\begin{aligned}
Z = E_f || E_p || E_a
\end{aligned}
\end{equation}
where the hidden dimension $d_h$ equals $3d_f+d_a$.

\subsection{Transformer and Regression Layer}
We apply vanilla transformers along temporal and spatial axes to capture intricate traffic relations. 
% In detail, we utilize the vanilla transformers by stacking spatio-temporal transformer layers.
\begin{table*}[]
\centering
\caption{Performance on METR-LA and PEMS-BAY.}
\small
\resizebox{\linewidth}{!}{
\begin{tabular}
{p{0.2cm}p{1.6cm}c|cccccccccccc}
\hline\hline
\multicolumn{2}{c}{Datasets}   & Metric & HI      & GWNet             & DCRNN   & AGCRN   & STGCN   & GTS     & MTGNN  & STNorm & GMAN  & PDFormer & STID & \textbf{STAEformer}  \\ \hline\hline
\multirow{9}{*}{\rotatebox{90}{METR-LA}}  & \multirow{3}{*}{\parbox{2cm}{\centering Horizon 3\\(15 min)} } & MAE    & 6.80    & 2.69            & 2.67    & 2.85    & 2.75      & 2.75    & 2.69   & 2.81 & 2.80    & 2.83    & 2.82     & \textbf{2.65}   \\
                          &                                                                                  & RMSE   & 14.21   & 5.15            & 5.16    & 5.53    & 5.29      & 5.27    & 5.16   & 5.57 & 5.55    & 5.45    & 5.53     & \textbf{5.11}   \\
                          &                                                                                  & MAPE   & 16.72\% & 6.99\%          & 6.86\%  & 7.63\%  & 7.10\%    & 7.12\%  & 6.89\% & 7.40\% & 7.41\%  & 7.77\%  & 7.75\%   & \textbf{6.85\%} \\
                          & \multirow{3}{*}{\parbox{2cm}{\centering Horizon 6\\(30 min)} }    & MAE    & 6.80    & 3.08            & 3.12    & 3.20    & 3.15      & 3.14    & 3.05   & 3.18 & 3.12    & 3.20    & 3.19     & \textbf{2.97}   \\
                          &                                                                                  & RMSE   & 14.21   & 6.20            & 6.27    & 6.52    & 6.35       & 6.33    & 6.13   & 6.59 & 6.49    & 6.46    & 6.57     & \textbf{6.00}   \\
                          &                                                                                  & MAPE   & 16.72\% & 8.47\%          & 8.42\%  & 9.00\%  & 8.62\%   & 8.62\%  & 8.16\% & 8.47\% & 8.73\%  & 9.19\%  & 9.39\%   & \textbf{8.13\%} \\
                          & \multirow{3}{*}{\parbox{2cm}{\centering Horizon 12\\(60 min)} }   & MAE    & 6.80    & 3.51            & 3.54    & 3.59    & 3.60       & 3.59    & 3.47   & 3.57 & 3.44    & 3.62    & 3.55     & \textbf{3.34}   \\
                          &                                                                                  & RMSE   & 14.20   & 7.28            & 7.47    & 7.45    & 7.43      & 7.44    & 7.21   & 7.51 & 7.35    & 7.47    & 7.55     & \textbf{7.02}   \\
                          &                                                                                  & MAPE   & 10.15\% & 9.96\%          & 10.32\% & 10.47\% & 10.35\%  & 10.25\% & 9.70\% & 10.24\% & 10.07\% & 10.91\% & 10.95\%  & \textbf{9.70\%} \\ \hline
\multirow{9}{*}{\rotatebox{90}{PEMS-BAY}} & \multirow{3}{*}{\parbox{2cm}{\centering Horizon 3\\(15 min)} } & MAE    & 3.06    & \textbf{1.30}   & 1.31    & 1.35    & 1.36       & 1.37    & 1.33   & 1.33 & 1.35    & 1.32    & 1.31     & 1.31            \\
                          &                                                                                  & RMSE       & 7.05    & \textbf{2.73}   & 2.76    & 2.88    & 2.88        & 2.92    & 2.80   & 2.82 & 2.90   & 2.83    & 2.79     & 2.78            \\
                          &                                                                                  &  MAPE      & 6.85\%  & \textbf{2.71\%} & 2.73\%  & 2.91\%  & 2.86\%   & 2.85\%  & 2.81\% & 2.76\% & 2.87\%  & 2.78\%  & 2.78\%   & 2.76\%          \\
                          & \multirow{3}{*}{\parbox{2cm}{\centering Horizon 6\\(30 min)} }    & MAE    & 3.06    & 1.63            & 1.65    & 1.67    & 1.70       & 1.72    & 1.66   & 1.65 & 1.65    & 1.64    & 1.64     & \textbf{1.62}   \\
                          &                                                                                  & RMSE       & 7.04    & 3.73            & 3.75    & 3.82    & 3.84        & 3.86    & 3.77   & 3.77 & 3.82   & 3.79    & 3.73     & \textbf{3.68}   \\
                          &                                                                                  & MAPE       & 6.84\%  & 3.73\%          & 3.71\%  & 3.81\%  & 3.79\%    & 3.88\%  & 3.75\% & 3.66\% & 3.74\%  & 3.71\%  & 3.73\%   & \textbf{3.62\%} \\
                          & \multirow{3}{*}{\parbox{2cm}{\centering Horizon 12\\(60 min)} }   & MAE    & 3.05    & 1.99            & 1.97    & 1.94    & 2.02        & 2.06    & 1.95   & 1.92 & 1.92    & 1.91    & 1.91     & \textbf{1.88}   \\
                          &                                                                                  & RMSE   & 7.03    & 4.60            & 4.60    & 4.50    & 4.63       & 4.60    & 4.50   & 4.45 & 4.49    & 4.43    & 4.42     & \textbf{4.34}   \\
                          &                                                                                  & MAPE   & 6.83\%  & 4.71\%          & 4.68\%  & 4.55\%  & 4.72\%    & 4.88\%  & 4.62\% & 4.46\% & 4.52\% & 4.51\%  & 4.55\%   & \textbf{4.41\%} \\ \hline\hline
\end{tabular}}
\label{table:metrla&pemsbay}
\end{table*}
Given hidden spatio-temporal representation $Z\in \mathbb{R} ^{T\times N\times d_h}$ with $T$ frames and $N$ spatial nodes, we obtain the query, key and value matrices through temporal transformer layers as: 
\begin{equation}
\begin{aligned}
Q^{(te)}=ZW_Q^{(te)}\ ,\ K^{(te)}=ZW_K^{(te)}\ ,\ V^{(te)}=ZW_V^{(te)}
\end{aligned}
\label{qkv}
\end{equation}
\noindent where $W_Q^{(te)},\ W_K^{(te)},\ W_V^{(te)}\ \in \mathbb{R} ^{d_h \times d_h}$ are learable parameters. Then we calculate the self-attention score as:
\begin{equation}
\begin{aligned}
A^{(te)}= Softmax\left(\frac{Q^{(te)}{K^{(te)}}^\top}{\sqrt{d_h}}\right)
\end{aligned}
\label{att_score}
\end{equation}
\noindent where $A^{(te)} \in \mathbb{R} ^{N\times T\times T}$ captures the temporal relations in different spatial nodes.
Finally, we obtain the output of temporal transformer $Z^{(te)}\in \mathbb{R} ^{T\times N\times d_h}$ as:
\begin{equation}
\begin{aligned}
Z^{(te)}= A^{(te)}V^{(te)}.
\end{aligned}
\label{value}
\end{equation}
Similarly, the spatial transformer layer performs as:
% \begin{align}
% &Q^{(sp)}=Z^{(te)}W_Q^{(sp)},K^{(sp)}=Z^{(te)}W_K^{(sp)},V^{(sp)}=Z^{(te)}W_V^{(sp)} \nonumber \\
% &A^{(sp)}=Softmax\left(\frac{Q^{(sp)}{K^{(sp)}}^\top}{\sqrt{d_h}}\right) \nonumber \\
% &Z^{(sp)}=A^{(sp)}V^{(sp)}
% \end{align}
\begin{equation}
    Z^{(sp)}=SelfAttention(Z^{(te)})
\end{equation}
where $SelfAttention$ follows Equation~\ref{qkv}, \ref{att_score}, \ref{value}, and $Z^{(sp)}\in \mathbb{R} ^{T\times N\times d_h}$ is the output of the spatial transformer. Notably, we also apply layer normalization, residual connection and multi-head mechanism.
% In summary, STAEformer utilizes vanilla transformers to capture intricate traffic patterns.

Finally, we leverage the output of the spatio-temporal transformer layers $Z'\in \mathbb{R} ^{T\times N\times d_h}$ to generate predictions. The regression layer can be formulated as:
\begin{equation}
\begin{aligned}
\hat{Y}= FC(Z')
\end{aligned}
\end{equation}
\noindent where $\hat{Y}\in \mathbb{R} ^{T'\times N\times d}$ is the prediction, $T'$ is the horizon of prediction. $d$ is the dimension of the output features which equals 1 in our case. Thus, the fully-connected layer regresses the dimension from $T\times d_h$ in $Z'$ to $T'\times(d=1)$ in $\hat{Y}$.

\begin{table}[b]
\small
\caption{Summary of Datasets.}
\begin{tabular}{cccc}
\hline\hline
\textbf{Dataset} & \textbf{\#Sensors (N)} & \textbf{\#Timesteps} & \textbf{Time Range} \\ \hline
METR-LA          & 207                    & 34,272               & 03/2012 - 06/2012   \\
PEMS-BAY         & 325                    & 52,116               & 01/2017 - 05/2017   \\
PEMS03          &  358 & 26,209 & 05/2012 - 07/2012 \\
PEMS04           & 307                    & 16,992               & 01/2018 - 02/2018   \\
PEMS07           & 883                    & 28,224               & 05/2017 - 08/2017   \\
PEMS08           & 170                    & 17,856               & 07/2016 - 08/2016   \\ \hline\hline
\end{tabular}
\label{table:datasets}
\end{table}

\begin{table}[t]
\small
\caption{Performance on PEMS03, 04, 07, and 08.}
\renewcommand\arraystretch{1.1}
\tabcolsep=0.7mm
\resizebox{8.5cm}{!}{
\begin{tabular}{c|ccc|ccc|ccc|ccc}
\hline\hline
Dataset                       & \multicolumn{3}{c|}{PEMS03} & \multicolumn{3}{c|}{PEMS04}                        & \multicolumn{3}{c|}{PEMS07}                       & \multicolumn{3}{c}{PEMS08}                        \\ \hline\hline
Metric      & MAE            & RMSE           & MAPE                  & MAE            & RMSE           & MAPE             & MAE            & RMSE           & MAPE            & MAE            & RMSE           & MAPE            \\ \hline\hline
HI               & 32.62 & 49.89 & 30.60\%            & 42.35          & 61.66          & 29.92\%          & 49.03          & 71.18          & 22.75\%         & 36.66          & 50.45          & 21.63\%         \\
GWNet                         & \textbf{14.59} & 25.24 & 15.52\% & 18.53          & \textbf{29.92}          & 12.89\%          & 20.47          & 33.47          & 8.61\%          & 14.40          & 23.39          & 9.21\%          \\
DCRNN    &  15.54 & 27.18 & 15.62\%                   & 19.63          & 31.26          & 13.59\%          & 21.16          & 34.14          & 9.02\%          & 15.22          & 24.17          & 10.21\%         \\
AGCRN      & 15.24 & 26.65 & 15.89\%                & 19.38          & 31.25          & 13.40\%          & 20.57          & 34.40          & 8.74\%          & 15.32          & 24.41          & 10.03\%         \\
STGCN    &  15.83 & 27.51 & 16.13\%                   & 19.57          & 31.38          & 13.44\%          & 21.74          & 35.27          & 9.24\%          & 16.08          & 25.39          & 10.60\%         \\
% StemGNN                       & 22.40          & 35.33          & 15.47\%          & 22.89          & 37.32          & 9.53\%          & 16.66          & 26.06          & 11.57\%         \\
GTS   &  15.41 & 26.15 &  15.39\%                    & 20.96          & 32.95          & 14.66\%          & 22.15          & 35.10          & 9.38\%          & 16.49          & 26.08          & 10.54\%         \\
MTGNN   & 14.85 & \textbf{25.23} & 14.55\%                    & 19.17          & 31.70          & 13.37\%          & 20.89          & 34.06          & 9.00\%          & 15.18          & 24.24          & 10.20\%         \\
STNorm    & 15.32 &25.93 & \textbf{14.37\%}                   & 18.96          & 30.98          & 12.69\%          & 20.50          & 34.66          & 8.75\%          & 15.41          & 24.77          & 9.76\%          \\
GMAN  & 16.87 & 27.92 & 18.23\%                       & 19.14          & 31.60          & 13.19\%          & 20.97          & 34.10          & 9.05\%          & 15.31          & 24.92          & 10.13\%         \\
PDFormer & 14.94 & 25.39 & 15.82\% & 18.36          & 30.03          & 12.00\% & 19.97          & 32.95          & 8.55\%          & 13.58          & 23.41          & 9.05\%          \\
STID    &  15.33 & 27.40 & 16.40\%                  & 18.38          & 29.95 & 12.04\%          & 19.61          & 32.79          & 8.30\%          & 14.21          & 23.28          & 9.27\%          \\
\textbf{STAEformer}   & 15.35 & 27.55 & 15.18\%            & \textbf{18.22} & 30.18          & \textbf{11.98}\%          & \textbf{19.14} & \textbf{32.60} & \textbf{8.01\%} & \textbf{13.46} & \textbf{23.25} & \textbf{8.88\%} \\ \hline\hline
\end{tabular}
}
\label{table:040708}
\end{table}

% \vspace{-0.3cm}
\section{Experiments}
\subsection{Experimental Setup}

\quad\,\textit{\textbf{Datasets.}} Our method is verified on six traffic forecasting benchmarks, i.e., METR-LA, PEMS-BAY, PEMS03, PEMS04, PEMS07, and PEMS08. The first two datasets were proposed by DCRNN~\cite{DCRNN}. The last four datasets were proposed by STSGCN ~\cite{STSGCN}. The time interval in the six datasets is 5 minutes, so there are 12 frames in each hour. More details are shown in Table~\ref{table:datasets}.
% The last four datasets were collected from California Performance of Transportation (PeMS) and proposed by STSGCN ~\cite{STSGCN}.

\textit{\textbf{Implementation.}} We implement the model with the PyTorch toolkit on a Linux server with a GeForce RTX 3090 GPU. METR-LA and PEMS-BAY are divided into the training, validation, and test sets in a fraction of 7:1:2. PEMS03, PEMS04, PEMS07 and PEMS08 are divided in a fraction of 6:2:2. In fact, the performance of our model is not sensitive to the hyper-parameters. For more details, the embedding dimension $d_f$ is 24 and the $d_a$ is 80. The number of layers $L$ is 3 for both spatial and temporal transformers. The number of heads is 4. We set the input and prediction length to be 1 hour, namely, $T$=$T'$=12. Adam is chosen as the optimizer with the learning rate decaying from 0.001, and the batch size is 16. We apply an early-stop mechanism if the validation error converges within 30 continuous steps. The code is available at \textbf{\url{https://github.com/XDZhelheim/STAEformer}}.

\begin{table}[t]
\caption{Ablation Study on PEMS04, PEMS07 and PEMS08.}
\renewcommand\arraystretch{1.3}
\tabcolsep=0.7mm
\resizebox{8.5cm}{!}{
\begin{tabular}{c|ccc|ccc|ccc}
\hline\hline
Dataset        & \multicolumn{3}{c|}{PEMS04}                        & \multicolumn{3}{c|}{PEMS07}                       & \multicolumn{3}{c}{PEMS08}                        \\ \hline\hline
Metric         & MAE            & RMSE           & MAPE             & MAE            & RMSE           & MAPE            & MAE            & RMSE           & MAPE            \\ \hline\hline
% w/o $Trans$   & 25.17          & 39.89          & 18.00\%          & 27.71          & 42.99          & 17.61\%         & 19.96          & 31.47          & 18.53\%         \\
w/o $E_a$      & 21.63          & 34.88          & 14.26\%          & 22.37          & 37.21          & 9.25\%          & 15.00          & 25.85        & 9.74\%          \\
w/o $E_p$      & 18.92          & 30.74          & 12.27\%          & 20.21          & 33.59          & 8.40\%          & 15.00          & 24.05          & 9.56\%          \\
% w/o $S$-$Trans$    & 18.51          & 30.16          & 12.11\%          & 19.57          & 33.16          & 8.19\%          & 13.49          & 22.99        & 8.92\%            \\
w/o $T$-$Trans$    & 18.50          & 30.34          & 12.24\%          & 19.87          & 33.67          & 8.29\%          & 13.83          & 23.79        & 9.09\%            \\
w/o $ST$-$Trans$   & 25.17          & 39.89          & 18.00\%          & 27.71          & 42.99          & 17.61\%         & 19.96          & 31.47          & 18.53\%         \\
\textbf{STAEformer} & \textbf{18.22} & \textbf{30.29} & \textbf{12.01\%} & \textbf{19.14} & \textbf{32.60} & \textbf{8.01\%} & \textbf{13.46} & \textbf{23.25} & \textbf{8.88\%} \\ \hline\hline
\end{tabular}
}
\label{table:ablation study}
\end{table}

\textit{\textbf{Metrics.}} We use three widely used metrics for traffic forecasting task, i.e, MAE, RMSE and MAPE. Following previous work, we select the average performance of all predicted 12 horizons on the PEMS03, PEMS04, PEMS07 and PEMS08 datasets. To evaluate the METR-LA and PEMS-BAY datasets, we compare the performance on horizon 3, 6 and 12 (15, 30, and 60 min).

\textit{\textbf{Baselines.}} 
% In this study, our proposed method is compared against several widely used baselines in the field. HI~\cite{HI} is a traditional model. GWNet~\cite{GWNet}, DCRNN~\cite{DCRNN}, AGCRN~\cite{AGCRN}, STGCN~\cite{STGCN}, StemGNN~\cite{StemGNN}, GTS~\cite{GTS} , and MTGNN~\cite{MTGNN} are typical STGNNs. Their used embeddings correspond to Figure~\ref{fig: model_embeddings}(a). STNorm~\cite{STNorm} aims to factorize traffic data. Note that there are several famous Transformer-based methods proposed for time series forecasting, including: Informer~\cite{Informer} and Pyra\-former~\cite{Pyraformer} focus on reducing the time and memory complexity; FEDformer~\cite{FEDformer} captures the global view of time series with linear complexity; Autoformer~\cite{Autoformer} adopts a decomposition architecture to tackle the intricate temporal patterns of the long-term future. However, none of these works is tailored for short-term traffic forecasting. Thus, we select two Transformer models that target the same task as ours, namely GMAN~\cite{GMAN} and PDFormer~\cite{PDFormer}, as our baselines, the input embeddings of which follow Figure~\ref{fig: model_embeddings}(b). STID~\cite{STID} with input embedding shown by Figure~\ref{fig: model_embeddings}(c), focuses on enhancing the spatio-temporal distinction in traffic data.
In this study, we compare our proposed method against several widely used baselines in the field. HI~\cite{HI} is a typical traditional model. We also consider STGNNs such as GWNet~\cite{GWNet}, DCRNN~\cite{DCRNN}, AGCRN~\cite{AGCRN}, STGCN~\cite{STGCN}, GTS~\cite{GTS}, and MTGNN~\cite{MTGNN}, which employ the embeddings shown in Figure~\ref{fig: model_embeddings}(a). Additionally, we examine STNorm~\cite{STNorm}, which focuses on factorizing traffic time series. While there exist Transformer-based methods for time series forecasting, such as Informer~\cite{Informer}, Pyraformer~\cite{Pyraformer}, FEDformer~\cite{FEDformer}, and Autoformer~\cite{Autoformer}, they are not specially tailored for short-term traffic forecasting. Hence, we select GMAN~\cite{GMAN} and PDFormer~\cite{PDFormer}, which are transformer models targeting the same task as ours. The input embeddings in ~\cite{GMAN} and ~\cite{PDFormer} follow the configuration in Figure~\ref{fig: model_embeddings}(b). Furthermore, we consider STID~\cite{STID}, which enhances the spaito-temporal distinction in traffic time series by utilizing the input embedding depicted in Figure~\ref{fig: model_embeddings}(c).

\begin{figure}[!t]
    \centering
    \includegraphics[width=\linewidth]{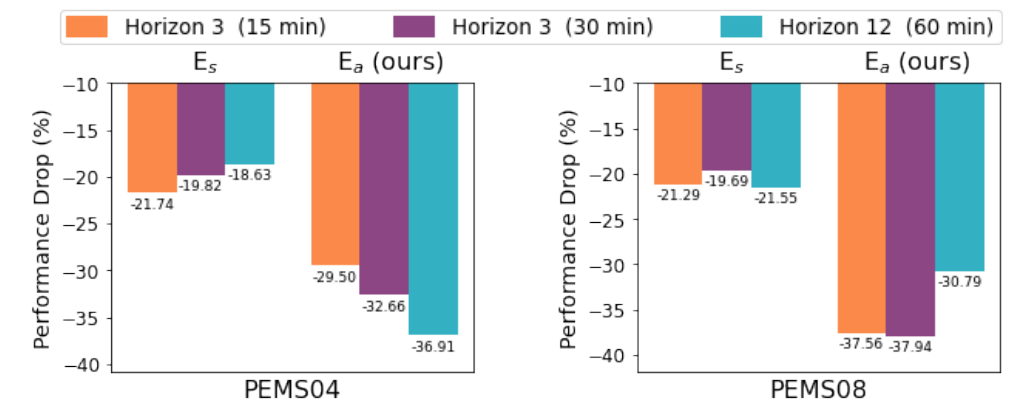}
    \caption{By shuffling the input along the temporal axis, our spatio-temporal adaptive embedding shows a more dramatic performance decrease than spatial embedding, indicating its capability in capturing chronological information.}
    \label{fig: performance_drop}
\end{figure}

% \vspace{-0.3cm}
\subsection{Performance Evaluation}

As shown in Table~\ref{table:metrla&pemsbay} and~\ref{table:040708}, our method achieves better performance on most metrics on all six datasets. STAEformer outperforms STGNNs by a large degree without any graph modeling. STNorm and STID also achieve competitive results, while Transformer-based models can better capture the intricate spatio-temporal relations. Compared with PDFormer, the encouraging results indicate STAEformer is a simpler but more effective solution.

% \vspace{-0.3cm}
\subsection{Ablation Study}\label{ablation study}

To evaluate the effectiveness of each part in STAEformer, we conduct ablation studies with 4 variants of our model as follows:
\begin{itemize}
    % \item 
    % \textbf{w/o $Trans$.} 
    % Transformer architecture is removed, only remaining a fully connected layer as the predictor.
    
    \item 
    \textbf{w/o $E_a$.}
    It removes spatio-temporal adaptive embedding $E_a$. 
    
    \item 
    \textbf{w/o $E_p$.}
    It removes periodicity embedding $E_p$, including day-of-week and timestamps-of-day embedding.
    
    \item 
    \textbf{w/o $T$-$Trans$.}
    It removes temporal transformer layers.

    \item 
    \textbf{w/o $ST$-$Trans$.}
    It removes both temporal transformer layers and spatial transformer layers.
    
    % \item 
    % \textbf{$STAEformer$.} 
    % It contains all components.
\end{itemize}
% The results from Table \ref{table:ablation study} demonstrate the effectiveness of the transformer architecture. Additionally, it is evident that different embeddings have an impact on the performance. periodicity embedding plays a crucial role in capturing daily and weekly patterns, while adaptive embedding is particularly effective in modeling spatio-temporal positional information for the transformer.
Table~\ref{table:ablation study} reveals the significance of various embeddings on the performance of our model. $E_p$ can capture daily and weekly patterns, while the proposed $E_a$ is vital for traffic modeling. Furthermore, the substantial performance degradation observed upon the removal of spatial or temporal transformer layers shows that our proposed embeddings effectively model the inherent spatio-temporal patterns in the traffic data. Therefore, both spatial and temporal layers are necessary for extracting those features.

\begin{figure}[!t]
    \centering
    \begin{subfigure}[t]{0.55\linewidth}
        \centering
        \includegraphics[width=\linewidth]{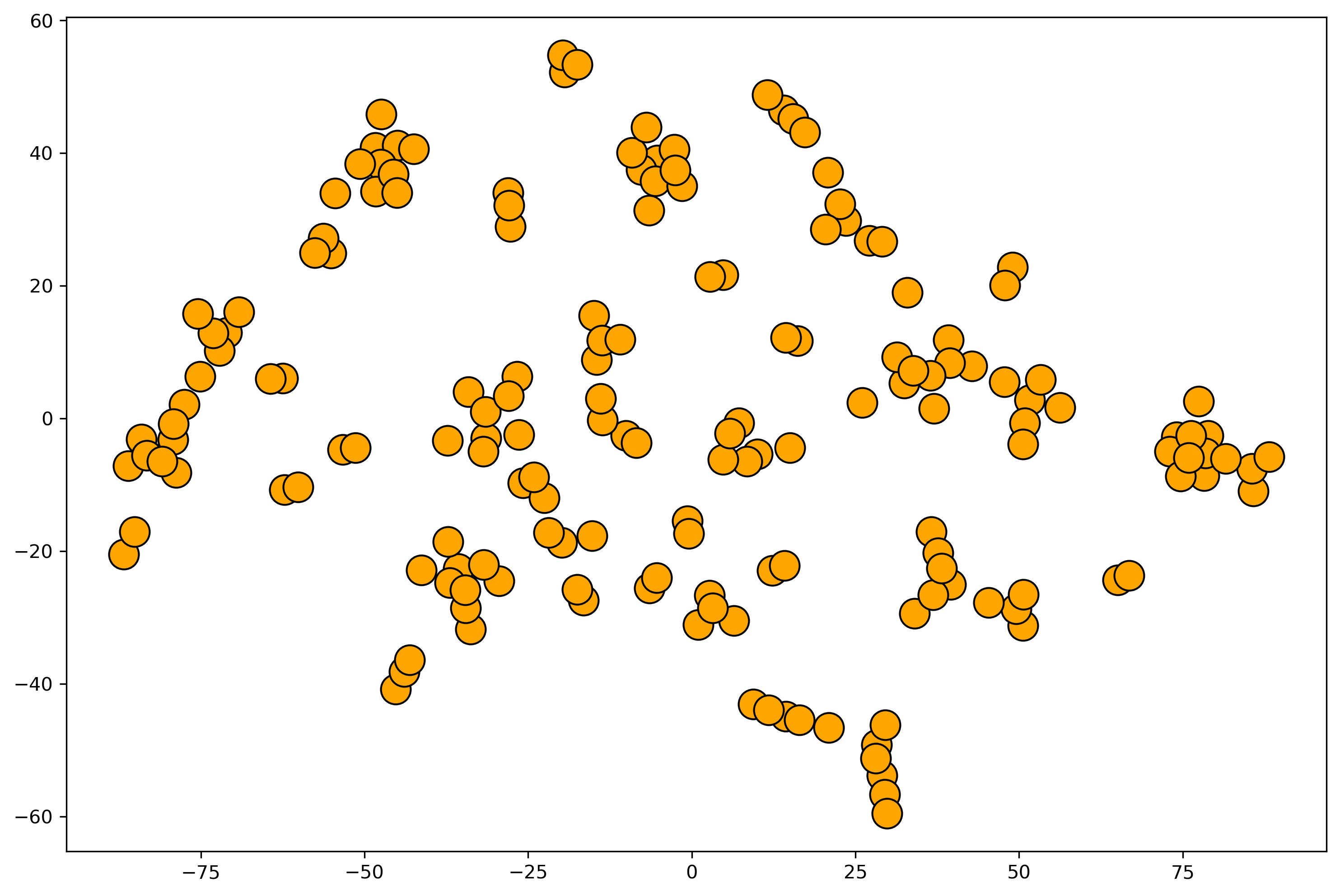}
        \caption{Spatial Axis}
        \label{fig: vis-s-emb}
    \end{subfigure}
    \begin{subfigure}[t]{0.44\linewidth}
        \centering
        \includegraphics[width=\linewidth]{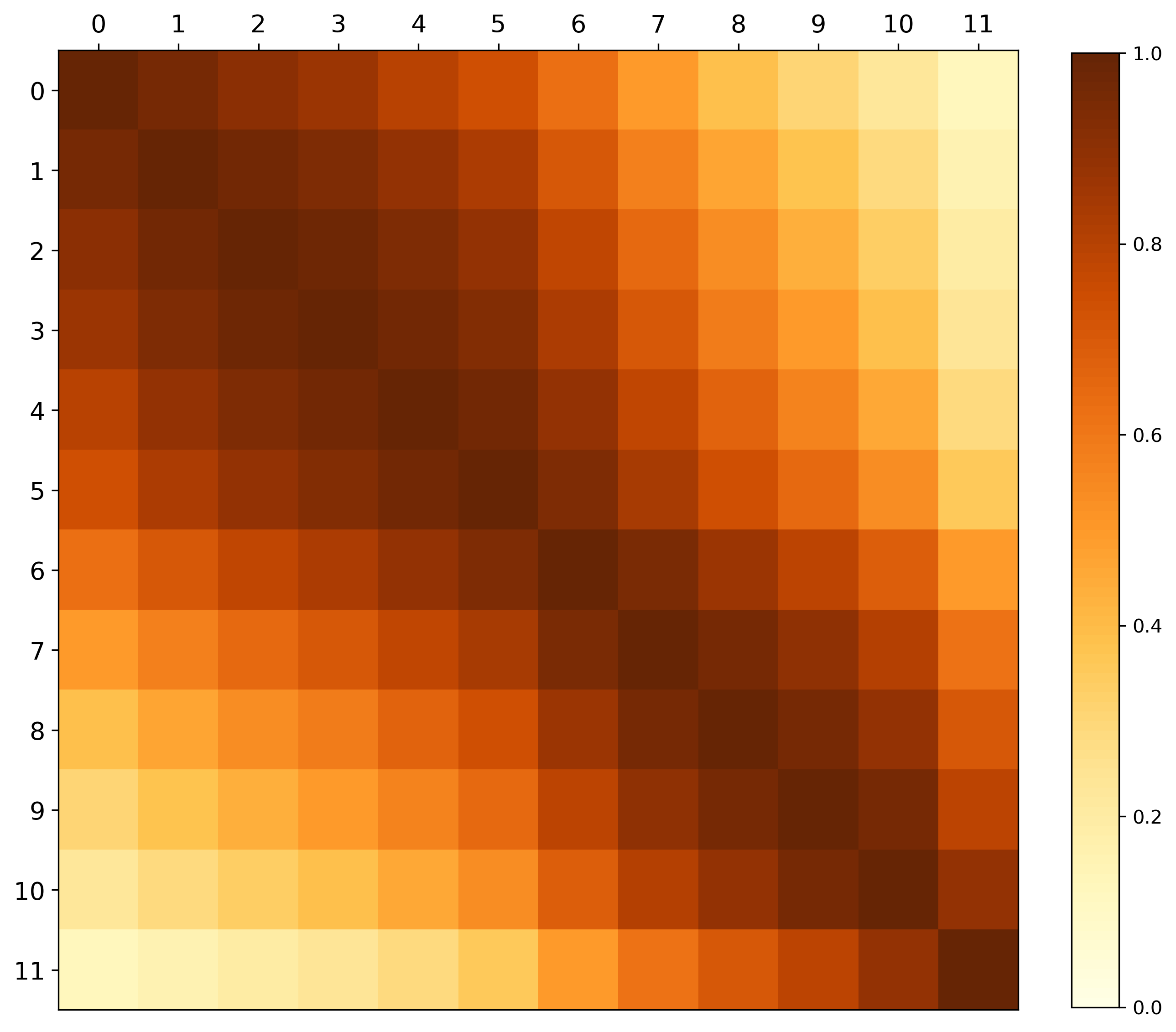}
        \caption{Temporal Axis}
        \label{fig: vis-t-emb}
    \end{subfigure}
    \caption{Visualization of Spatio-Temporal Adaptive Embedding $E_a$ on PEMS08.}
    \label{fig: vis}
\end{figure}

% \vspace{-0.3cm}
\subsection{Case Study}
% \noindent\textbf{Chronological Information Capturing.}
\quad\,\textit{\textbf{Comparison with Spatial Embedding.}} To validate the effectiveness of spatio-temporal adaptive embedding in capturing chronological order information implied in the $T$ input frames, we conduct more experiments on PEMS04 and PEMS08 datasets. Following~\cite{AreTrans}, we randomly shuffle the raw input along temporal axis $T$. For comparison, we replace our spatio-temporal adaptive embedding $E_a$ with spatial embedding $E_s$ that was used in~\cite{STID,PDFormer}. As shown by Figure~\ref{fig: performance_drop}, our model has more severe performance degradation when shuffling the raw input along temporal axis $T$. It means that spatio-temporal adaptive embedding $E_a$ makes our model more sensitive to the chronological order, while the model with $E_s$ is relatively insensitive. In summary, $E_a$ can better model the chronological information in the raw input and other intricate traffic patterns, which is crucial to the task.

\textit{\textbf{Visualization of Spatio-Temporal Adaptive Embedding.}} Figure~\ref{fig: vis} further provides visualizations of our proposed spatio-temporal embedding $E_a$ on the spatial and temporal axes by taking PEMS08 dataset as an example. For the spatial axis, we use t-SNE~\cite{TSNE} to get Figure~\ref{fig: vis-s-emb}. It shows that the embeddings of different nodes naturally form into clusters, which matches the spatial characteristics of the traffic data. For the temporal axis, we calculate the correlation coefficient across the 12 input frames and draw a heat map as Figure~\ref{fig: vis-t-emb}. It shows that each frame is highly correlated to the nearby frames, and the correlation gradually decreases for further frames. This shows that our proposed $E_a$ models the chronological information in the time series correctly.

% TODO
% 1. 完成vis
% 2. 改query那一段
% 3. w/o s-att, w/o t-att
% 4. linear predictor

% w-LSE
% Shuf. 不需要简写了
% case study 的 subsubsec -> textit 和上面统一

% \vspace{-0.3cm}
\section{Conclusion}
% In this paper, we focus on representation techniques rather than complicated models. Our proposed STAEN combines a novel spatio-temporal adaptive embedding with simple vanilla transformers. Despite its concise structure, STAEN achieves better performance. The experiments demonstrate that our model effectively captures intrinsic spatio-temporal dependencies. By leveraging the power of input embedding, our approach offers a promising solution for addressing the challenges in traffic forecasting.

In this study, we focus on a basic representation learning technique for traffic time series forecasting, i.e., input embedding. We propose a novel spatio-temporal adaptive embedding that can work on vanilla transformers to achieve the SOTA performance on six traffic benchmarks. Further studies demonstrate that our model can effectively capture intrinsic spatio-temporal dependencies. Instead of designing complicated models, our study shows a promising direction for addressing the challenges in traffic forecasting.

%The further studies demonstrate that our model can effectively capture intrinsic spatio-temporal dependencies.
% the further studies 概括了 section 4.3 & 4.4.
% studies/analyses 这块想说的是 各种可解释性的分析 其实是想说进一步的分析。

\begin{acks}
This work was partially supported by the grants of National Key Research and Development Program of China (2021YFB1714400) .
\end{acks}

\bibliographystyle{ACM-Reference-Format}
\bibliography{reference}

\end{document}